\documentclass[lettersize,journal]{IEEEtran}
\usepackage{amsmath,amsfonts}
\usepackage{algorithmic}
\usepackage{algorithm}
\usepackage{array}
\usepackage[caption=false,font=normalsize,labelfont=sf,textfont=sf]{subfig}
\usepackage{textcomp}
\usepackage{stfloats}
\usepackage{url}
\usepackage{verbatim}
\usepackage{graphicx}
\usepackage{cite}
\usepackage{soul}
\usepackage{xcolor}
\usepackage{xspace}

\definecolor{hlblue}{rgb}{0.8, 0.85, 1} 

\begin{document}

\title{Visual Heading Prediction for Autonomous Aerial Vehicles}

\author{Reza Ahmari$^{1}$, Ahmad Mohammadi$^{2}$, Vahid Hemmati$^{2}$, Mohammed Mynuddin$^{2}$, Parham Kebria$^{2}$,\\
Mahmoud Nabil Mahmoud$^{2}$, Xiaohong Yuan$^{1}$, Abdollah Homaifar$^{2*}$
\thanks{$^{1}$ Authors are with the Department of Computer Science at North Carolina A\&T State University, Greensboro, NC 27411, USA.}
\thanks{$^{2}$ Authors are with the Department of Electrical and Computer Engineering at North Carolina A\&T State University, Greensboro, NC 27411, USA. (*Corresponding author: homaifar@ncat.edu)}
}

\markboth{IEEE TRANSACTION ON INTELLIGENT VEHICLES,~Vol.~xx, No.~x, August~2025}
{Ahmari \MakeLowercase{\textit{et al.}}: Vision-Based Heading Prediction for Autonomous Aerial Vehicles via Neural Architecture}


\maketitle

\begin{abstract}
The integration of Unmanned Aerial Vehicles (UAVs) and Unmanned Ground Vehicles (UGVs) is increasingly central to the development of intelligent autonomous systems for applications such as search and rescue, environmental monitoring, and logistics. However, precise coordination between these platforms in real-time scenarios presents major challenges, particularly when external localization infrastructure such as GPS or GNSS is unavailable or degraded\cite{shi2018}. This paper proposes a vision-based, data-driven framework for real-time UAV-UGV integration, with a focus on robust UGV detection and heading angle prediction for navigation and coordination. The system employs a fine-tuned YOLOv5 model to detect UGVs and extract bounding box features, which are then used by a lightweight artificial neural network (ANN) to estimate the UAV’s required heading angle. A VICON motion capture system was used to generate ground-truth data during training, resulting in a dataset of over 13,000 annotated images collected in a controlled lab environment. The trained ANN achieves a mean absolute error of 0.1506° and a root mean squared error of 0.1957°, offering accurate heading angle predictions using only monocular camera inputs. Experimental evaluations achieve 95\% accuracy in UGV detection. This work contributes a vision-based, infrastructure-independent solution that demonstrates strong potential for deployment in GPS/GNSS-denied environments, supporting reliable multi-agent coordination under realistic dynamic conditions. A demonstration video showcasing the system’s real-time performance, including UGV detection, heading angle prediction, and UAV alignment under dynamic conditions, is available at: \url{https://github.com/Kooroshraf/UAV-UGV-Integration}
\end{abstract}


\section{Introduction}
\IEEEPARstart{T}{he} integration of Unmanned Aerial Vehicles (UAVs) and Unmanned Ground Vehicles (UGVs) has emerged as a powerful paradigm in multi-agent systems, offering significant advantages for surveillance, search and rescue, precision agriculture, and autonomous logistics \cite{munasinghe2024comprehensive}. UAVs provide agility and a wide field of view, while UGVs offer stable ground-level interaction and payload capacity. Their coordinated deployment enables collaborative behaviors such as aerial monitoring, dynamic path planning, and high-precision landing in both structured and unstructured environments \cite{ccacska2014survey}. Earlier approaches like market-based multi-robot task allocation also explored decentralized collaboration, especially in logistics and exploration scenarios \cite{zlot2014multirobot}. More recently, graph-based deep learning frameworks have been introduced for dynamic UAV–UGV task allocation in complex environments, such as the Adaptive Depth Graph Neural Network (ADGNN) model, which enhances decision-making efficiency and adaptability \cite{10670550}.

Despite recent progress, achieving reliable UAV-UGV integration in real-time remains a challenging task, particularly in unstructured terrain requiring fast and accurate trajectory adaptation \cite{xiao2021learning, huang2022survey}. Key technical hurdles include accurate heading angle estimation for alignment, robust detection and tracking of UGVs in dynamic scenes, and secure, real-time decision-making under limited sensor availability. Many existing approaches rely on high-cost sensor fusion, such as GPS-INS, LiDAR, or motion capture systems, which complicate deployment in real-world scenarios \cite{merriaux2017study, xin2022vision}. These solutions, although precise, are impractical in GPS-denied or infrastructure-sparse environments. Similar concerns about dataset bias and scalability have been raised in broader autonomous driving contexts, where dataset coverage and annotation quality directly impact generalization in deployment \cite{liu2024survey}.

More broadly, this research contributes to the development of adaptive AI-based perception frameworks for UAV-UGV operations in conditions where conventional localization infrastructure is unavailable or degraded. While our experiments were conducted indoors using a VICON motion capture system, the proposed approach serves as a baseline for future adaptation to more complex settings, such as outdoor or semi-structured environments where illumination changes, partial occlusions, and temporary localization outages may occur. These conditions highlight the need for vision-based systems that operate independently of GPS or GNSS. Recent studies on GPS spoofing detection using time series analysis and data-driven anomaly detection \cite{ahmadvehicular} reinforce the importance of localization-independent perception. Although the present work does not address adversarial threats or contested airspace directly, it is motivated by the broader goal of enabling robust navigation in degraded or infrastructure-limited environments. This aligns with priorities outlined by the Department of Transportation, NASA, and the Department of Homeland Security, and reflects ongoing trends toward learning-based perception for unstructured environments \cite{guastella2020learning}.

Recent advances in computer vision and deep learning have introduced new possibilities for lightweight and cost-effective navigation systems. A comprehensive survey on vision-based UAV navigation emphasizes the growing use of monocular and deep-learning-based methods for perception and autonomy \cite{lu2018survey}. YOLO-based object detectors have demonstrated strong performance in real-time object identification tasks \cite{redmon2018, chowdhury2024}, while neural network regressors offer a viable solution for estimating spatial orientation from image-derived features. By leveraging these technologies, vision-only architectures can now offer heading angle prediction and object localization with reasonable accuracy and speed.

In this work, we propose a data-driven framework that eliminates the need for external localization systems by relying solely on onboard camera inputs. The system integrates a fine-tuned YOLOv5 model for UGV detection with a compact ANN to predict the UAV's heading angle from bounding box features. During training and evaluation, we employed a VICON motion capture system to generate high-fidelity ground truth for over 13,000 annotated images, enabling robust model supervision \cite{merriaux2017study}. The trained ANN was able to accurately predict heading angles with a root mean squared error of 0.1957°, and the system as a whole achieved a 95\% detection accuracy.

This research delivers a deployable and scalable solution for UAV–UGV coordination that is sensor-independent, computation-efficient, and capable of operating in complex, dynamic environments. By demonstrating that vision-only systems can achieve high-precision autonomous behavior, it represents a significant step toward practical real-world deployment of cooperative aerial–ground robotic systems. The proposed framework advances the state of vision-based UAV–UGV coordination by introducing a lightweight, markerless architecture for heading angle prediction that removes the need for external localization during deployment. While training and validation were conducted in a controlled indoor environment with VICON supervision, the design emphasizes computational efficiency, adaptability, and readiness for extension to GPS/GNSS-denied and infrastructure-sparse scenarios.

The main contributions of this work are as follows:
\begin{itemize}
    \item \textbf{A vision-only heading prediction framework} that integrates a fine-tuned YOLOv5 detector with a compact ANN regressor, enabling real-time UAV alignment with UGVs using only monocular bounding-box features.
    \item \textbf{A custom, high-fidelity dataset} comprising over 13{,}000 annotated images synchronized with precise VICON-based ground truth, capturing diverse UGV trajectories for robust model training.
    \item \textbf{Demonstration of sub-degree heading accuracy} (MAE = 0.1506°, RMSE = 0.1957°) and 95\% UGV detection precision, achieved with a computation-efficient model suitable for embedded platforms.
    \item \textbf{Closed-loop, real-time evaluation} confirming system stability, low latency (31 ms/frame), and suitability for GPS-denied operation without reliance on external infrastructure during runtime.
\end{itemize}

\section{Related Work}

The integration of UAVs and UGVs has been a focal point in autonomous systems research due to its potential in enabling multi-agent coordination across various domains, including environmental monitoring, military surveillance, and disaster response \cite{ccacska2014survey}. Traditionally, UAVs and UGVs have operated independently, with limited collaboration. Recent studies have emphasized the advantages of shared autonomy, where UAVs provide top-down situational awareness while UGVs interact with the environment on the ground, forming a cooperative feedback loop.

Early approaches to coordination relied heavily on GPS and inertial navigation systems (INS) for position estimation and control. However, such solutions often fail in GPS-denied or occluded environments, motivating the development of vision-based alternatives \cite{xin2022vision}. Marker-based landing strategies, such as those using fiducial markers or predefined patterns, have shown success in controlled environments but lack robustness in dynamic or unstructured scenes \cite{yu2018deep}. Motion capture systems like VICON provide highly accurate ground-truth data, yet their reliance on fixed infrastructure and limited scalability makes them impractical for real-world deployment \cite{merriaux2017study}. In contrast, methods like visual teach-and-repeat offer infrastructure-free autonomy by using onboard vision for route memorization and long-range path repetition \cite{furgale2010visual}.

To address these limitations, deep learning methods have been increasingly adopted. Object detection models such as YOLO have demonstrated real-time performance and high precision in detecting targets such as UGVs, pedestrians, and other objects in cluttered environments \cite{redmon2018, chowdhury2024}. YOLOv5, in particular, has been widely used for edge-deployable applications due to its balance between accuracy and inference speed \cite{zhang2022real}. These models output bounding box parameters that can serve not only for detection but also for spatial reasoning tasks, such as heading angle estimation.

Prior work has explored the use of regression models to estimate spatial relationships between aerial and ground agents. In this context, neural networks trained on visual features have been proposed to predict relative orientation or distance without explicit geometric modeling. Such approaches reduce reliance on depth sensors or external localization infrastructure and are suitable for embedded applications. Some monocular systems also leverage dense 3D flow and visual odometry for spatial awareness without depth sensors \cite{zhao2018learning}.

\begin{figure*}[!hbt]
    \centering
    \includegraphics[width=0.9\textwidth]{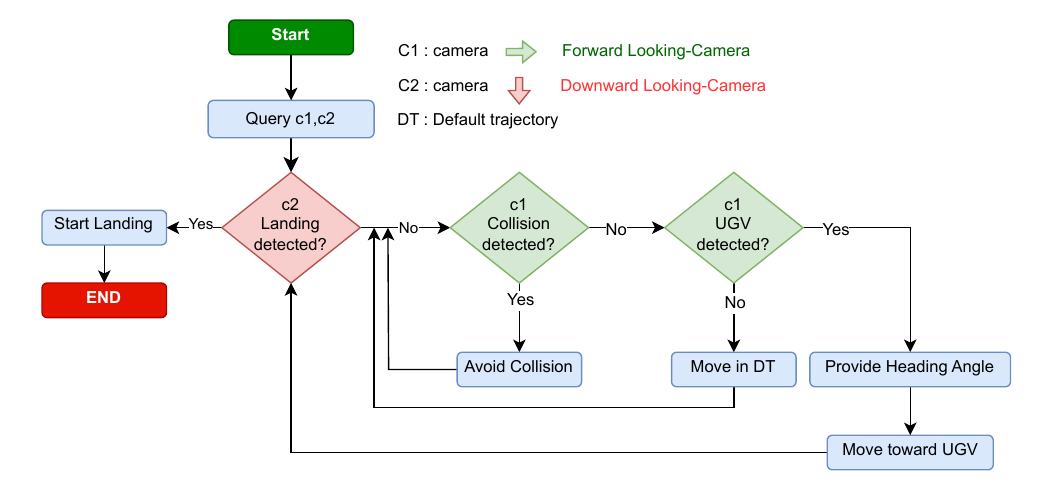}
    \caption{Overview of the proposed UAV-UGV coordination framework. The forward-facing camera (C1) is used for UGV detection and heading estimation. The ANN predicts alignment. The downward-facing camera (C2) confirms landing.}
    \label{fig:system_diagram}
\end{figure*}

Collision detection is another critical component in autonomous UAV systems. The DroNet model introduced by Loquercio et al. \cite{loquercio2018} exemplifies how deep convolutional networks can learn safe navigation policies from driving data. Although primarily used for obstacle avoidance, such vision-based models highlight the growing trend toward using end-to-end learning systems in robotics.

Security remains a growing concern in vision-based navigation. Recent studies have shown how deep learning models are susceptible to Trojan attacks, where hidden triggers embedded during training can hijack model behavior \cite{trojan2023}. These findings underscore the need for secure, interpretable, and robust model architectures, especially when deployed in safety-critical systems like UAV-UGV operations.

In summary, existing literature supports the feasibility of deep learning-based UAV-UGV coordination and highlights a broader trend toward learning-based perception and navigation across autonomous systems \cite{tang2022perception}. However, a clear gap remains in frameworks that unify object detection, heading angle prediction, and landing guidance into a lightweight, sensor-independent solution suitable for real-time deployment. This work addresses that gap by leveraging YOLOv5 and an ANN trained solely on visual bounding box features to enable robust coordination and landing in dynamic environments.

\section{Methodology}
The proposed framework consists of three main modules: (1) vision-based UGV detection using YOLOv5, (2) heading angle calculation using ground truth data from a motion capture system during training, and (3) heading angle prediction using an ANN trained on bounding box features. The system enables \textit{real-time} UAV-UGV coordination by combining object detection, heading prediction, collision avoidance, and control. It is designed for GPS-denied environments using monocular camera input, lightweight neural models, and real-time execution within a Robot Operating System (ROS) network.

\subsection{Heading Angle Estimation: Problem Formulation and Learning Objective}

The core objective of this work is to develop a vision-based model that estimates the relative heading angle between the UAV and UGV using only visual features, without relying on external localization at runtime.

Let \( \mathbf{z}_{cam}(t) \in \mathbb{R}^d \) denote the vector of visual features extracted at time \( t \) from the bounding box detected by YoloV5. Let \( \theta_{true}(t) \) denote the ground-truth heading angle derived from VICON positional data at the same timestamp.
The goal is to learn a function \( f: \mathbb{R}^d \rightarrow \mathbb{R} \) such that:

\[
\hat{\theta}(t) = f(\mathbf{z}_{cam}(t))
\]

where:
\begin{itemize}
    \item \( \hat{\theta}(t) \) is the predicted heading angle,
    \item \( f \) is a data-driven model (e.g., a neural network),
    \item \( \mathbf{z}_{cam}(t) = [c_x, c_y, A, \alpha] \) includes normalized center coordinates, bounding box area, and aspect ratio.
\end{itemize}
These features were selected because they provide a compact yet informative representation of the UGV’s relative position and scale in the image. The center coordinates encode the spatial alignment between the UAV and UGV, while the area and aspect ratio implicitly capture distance and orientation changes. Compared to alternatives such as diagonal length or full bounding box coordinates, these features are less sensitive to perspective distortion and redundant correlations, making them more robust under partial occlusion and varying viewing angles.

\subsubsection*{Model Training Objective}

The model \( f \) is trained to minimize the mean squared error (MSE) between predicted and true heading angles across all \( N \) samples:

\[
L(f) = \frac{1}{N} \sum_{t=1}^{N} \left( f(\mathbf{z}_{cam}(t)) - \theta_{true}(t) \right)^2
\]

This formulation enables learning a compact regression model capable of real-time heading estimation using monocular camera input alone.

\subsection{Framework Components}

The overall system comprises the following main components:

\begin{itemize}
    \item \textbf{UGV Detection:} A YOLOv5 model, fine-tuned on a custom grayscale dataset, detects UGVs in real time. The predicted bounding boxes serve as input for both spatial reasoning and heading prediction \cite{redmon2018, chowdhury2024}.
    
    \item \textbf{Heading Angle Prediction:} An ANN consumes normalized bounding box features (center coordinates, area, and aspect ratio) to predict the UAV’s desired heading angle. This replaces geometric computations with a data-driven approach and allows operation without external localization systems.
    
    \item \textbf{Collision Avoidance:} The DroNet model \cite{loquercio2018} is used for forward obstacle detection. When potential collisions are identified via camera C1, the UAV takes evasive action before resuming UGV alignment.
    
    \item \textbf{Landing Confirmation:} Visual cues from camera C2 confirm pad presence once alignment is achieved. A comprehensive evaluation of the landing procedure, including robustness and security aspects, is provided in \cite{ahmari2025smc}.

\end{itemize}

A high-level schematic of the full pipeline, including sensor inputs, deep learning components, and system control, is illustrated in Figure~\ref{fig:system_diagram}.

The modularity of the architecture supports plug-and-play operation, allowing easy replacement of individual models or sensors, making the framework suitable for real-world deployment in resource-constrained and infrastructure-free environments.

\section{Dataset Collection and Processing}

To train and validate the deep learning components of our system, we constructed a custom dataset in an indoor experimental setup using VICON as the localization module for training purposes, synchronized video streams, and UGV mobility patterns. The dataset includes over 13{,}000 images capturing UGV movement from the UAV’s forward-facing camera, along with precise heading angle labels derived from spatial coordinates.

\subsection{Hardware Setup and Sensor Configuration}

The UAV platform is equipped with two monocular cameras: a forward-facing camera (\textbf{C1}) for object detection, heading estimation, and collision monitoring, and a downward-facing camera (\textbf{C2}) used for confirming safe landing once the UGV is aligned. The UGV is tracked using motion capture during training but operates independently during deployment. All processing is conducted on an embedded platform with real-time control loops interfaced via ROS.

For dataset generation and offline evaluation, a VICON motion capture system is used. This system provides high-precision six degrees-of-freedom (6DOF) positional data for both UAV and UGV, enabling accurate annotation of visual frames and heading angles \cite{merriaux2017study}. Over 13,000 frames were collected using synchronized camera streams and positional logs, creating a dataset with bounding box annotations, spatial metadata, and temporal alignment.

\subsection{Software and Control Flow}

The system employs a hybrid Windows-Linux software environment to integrate the VICON Tracker (Windows-based) with the Linux-based ROS control stack. A custom \textbf{Python API} was developed to extract and synchronize data in real time. Using the Windows Subsystem for Linux (WSL), we eliminate the need for dual-machine configurations, ensuring efficient data flow across the entire system.

The ROS architecture follows a master-slave model where the UAV acts as the master node and the UGV as a slave. Command and feedback messages are exchanged over dedicated topics, including velocity commands (\texttt{/cmd\_vel}) and sensor updates. This structure allows real-time maneuvering of the UAV in response to visual inputs and model predictions.

\subsection{Data Acquisition Setup}

\begin{figure}[h]
    \centering
    \includegraphics[width=0.41\textwidth]{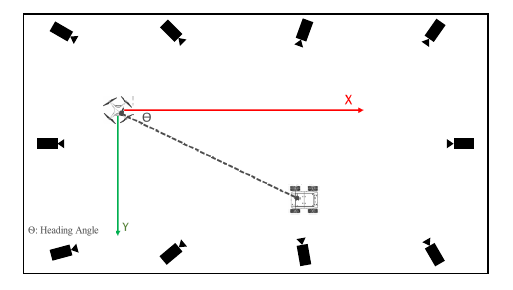}
    \caption{Top view of the laboratory scene showing VICON camera placement and UGV/UAV layout.}
    \label{fig:lab_topview}
\end{figure}

Data was collected in a motion capture arena equipped with ten infrared VICON cameras, providing full 6DOF tracking of both the UAV and UGV, as illustrated in Figure~\ref{fig:lab_topview}.
During data collection, the UAV remained stationary while a single UGV navigated through the scene. The primary objective was to learn the relative position and orientation of the UGV with respect to the UAV’s camera frame; therefore, all training images were captured in UAV-centric coordinates. This approach ensured consistent alignment between visual observations and ground-truth data, which is essential for accurate heading angle estimation. By emphasizing the relative UGV pose rather than UAV motion dynamics, the dataset effectively captured the most informative features for heading angle prediction and UGV detection. Simultaneously, the VICON system recorded the real-time positions and orientations of both vehicles for precise ground-truth generation. A custom Python API was developed to synchronize the image timestamps with the corresponding positional data. This enabled frame-level alignment of UGV position relative to the UAV's viewpoint \cite{merriaux2017study}.

\subsection{VICON Tracker Visualization}

\begin{figure}[h]
    \centering
    \includegraphics[width=0.5\textwidth]{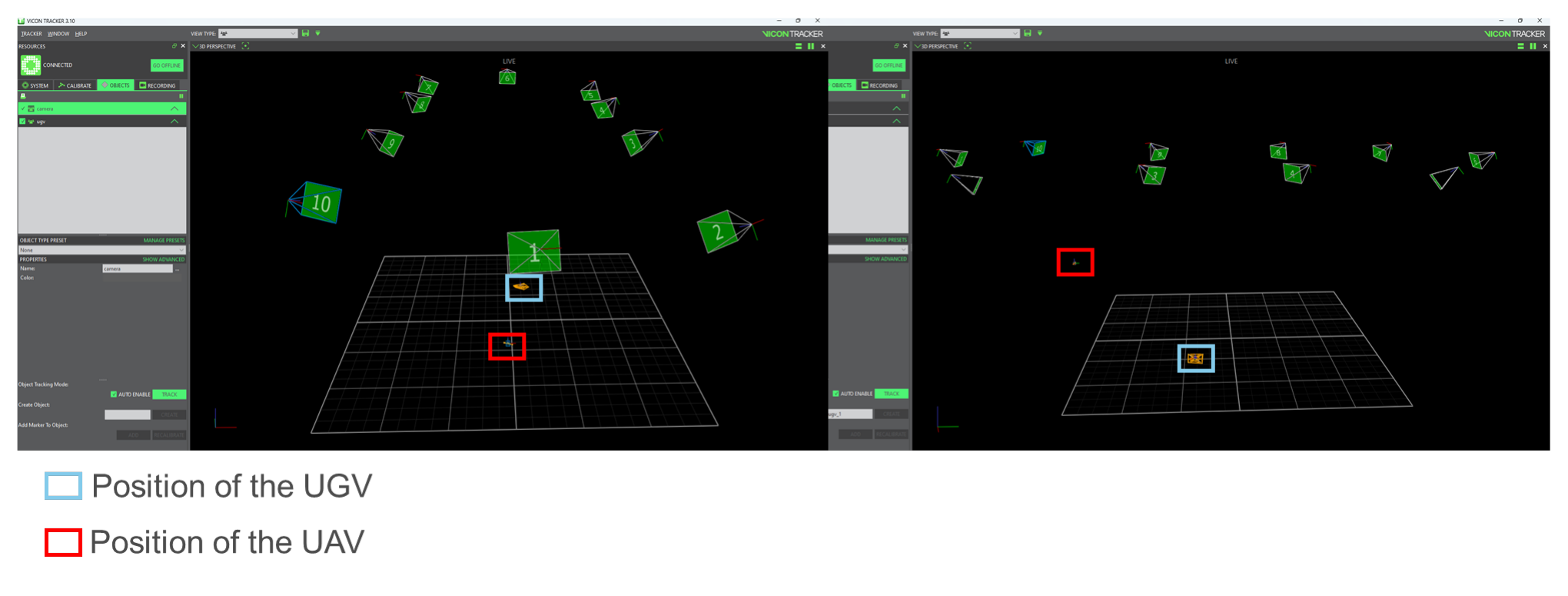}
    \caption{VICON Tracker interface showing UAV and UGV marker positions in the 3D capture space. The left panel presents a top-down (horizontal) view of the tracking volume, while the right panel shows a side (elevation) perspective, enabling spatial verification of relative height and position. The red and blue bounding boxes highlight the tracked UAV and UGV, respectively.}
    \label{fig:vicon_tracker}
\end{figure}

The positions of both UAV and UGV were visualized in real time using the VICON Tracker software interface. Each vehicle was marked with a distinct set of reflective markers, enabling the VICON system to track their full 6DOF motion during dataset generation. An example of the tracking interface is shown in Figure~\ref{fig:vicon_tracker}.

\subsection{Data Annotation and Preprocessing}

\begin{figure}[h]
    \centering
    \includegraphics[width=0.5\textwidth]{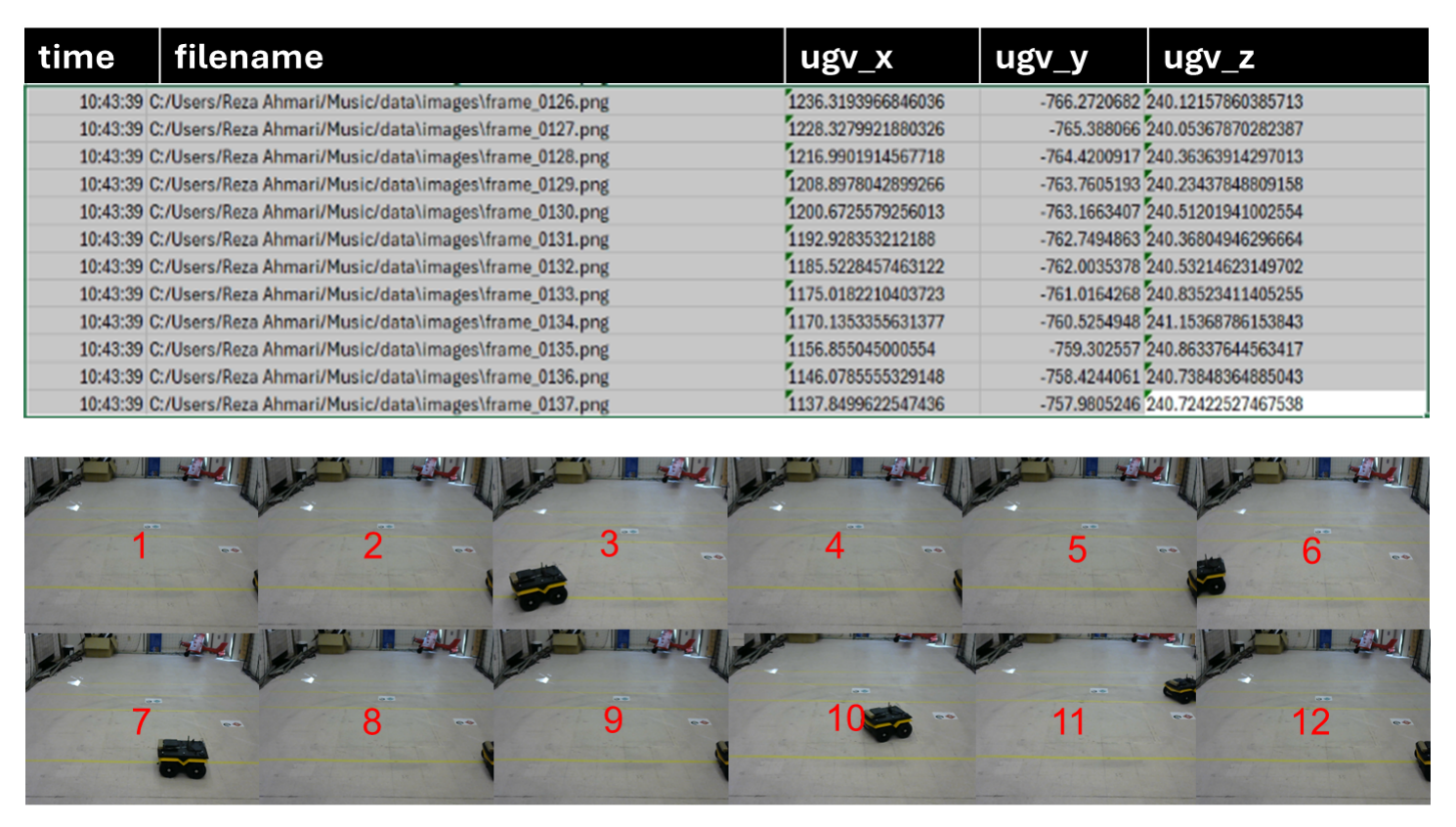}
    \caption{Data generation pipeline showing raw camera frames (bottom) aligned with VICON-recorded positional metadata (top).}
    \label{fig:data_generation}
\end{figure}

Each image was manually annotated with a bounding box around the UGV using label-editing tools. This manual annotation was necessary because YOLOv5 does not include a predefined UGV class, requiring custom labeling to train the model for our specific use case. Such annotation practices reflect broader challenges noted in autonomous driving datasets, where annotation quality and labeling consistency are key factors influencing downstream perception performance \cite{liu2024survey}. Figure~\ref{fig:data_generation} illustrates a sequence of collected frames and corresponding ground truth UGV positions as logged by the VICON system, showing how visual data and 3D labels were synchronized.

To streamline computation and enhance generalization, the images were converted to grayscale and resized to $640 \times 640$ resolution before training. After filtering out blurry, static, or duplicate frames, 9{,}000 high-quality annotated images were retained for further use.
To expand the dataset diversity, we applied several augmentation techniques including horizontal and vertical flips, random rotations, brightness adjustments, and minor translations. This step was critical for training YOLOv5 to detect UGVs under varied conditions, such as partial occlusion and different lighting patterns.

\subsection{Ground Truth Generation}

The heading angle for each image frame was calculated using the positional difference between the UAV and UGV recorded by the VICON system. Given the UAV position $(x_{\text{uav}}, y_{\text{uav}})$ and the UGV position $(x_{\text{ugv}}, y_{\text{ugv}})$, the ground truth heading angle $\theta_{true}$ was derived as:
\begin{equation}
    \theta_{true} = \text{atan2}(y_{\text{ugv}} - y_{\text{uav}},\ x_{\text{ugv}} - x_{\text{uav}})
\end{equation}

\subsection{Feature Extraction for ANN Training}

After YOLOv5 was trained and deployed on the annotated dataset, Each detected UGV in an image frame produces bounding box coordinates $(x_1, y_1, x_2, y_2)$, generated by the YOLOv5 detector. These coordinates are normalized with respect to the image resolution (640 × 640 pixels) to derive four features that serve as the input vector to the ANN:

\begin{equation}
c_x = \frac{x_1 + x_2}{2 \times 640}, \quad
c_y = \frac{y_1 + y_2}{2 \times 640},
\end{equation}
\begin{equation}
A = \frac{(x_2 - x_1)(y_2 - y_1)}{640 \times 640}, \quad
\alpha = \frac{y_2 - y_1}{x_2 - x_1}
\end{equation}

Where:
\begin{itemize}
    \item $c_x, c_y$: normalized center coordinates of the bounding box  
    \item $A$: normalized area of the bounding box  
    \item $\alpha$: aspect ratio of the bounding box  
\end{itemize}

These features were chosen for their ability to capture the relative position, scale, and geometric proportions of the UGV, which are crucial for heading angle estimation. The dataset was split into training (80\%), validation (10\%), and testing (10\%) subsets to ensure model generalization and unbiased evaluation.

\section{YOLOv5-Based UGV Detection and Performance}

To enable robust and real-time UGV detection, we fine-tuned a YOLOv5 model on the preprocessed dataset. YOLOv5 was chosen for its strong balance between inference speed and accuracy, especially in embedded systems where lightweight models are preferred \cite{zhang2022real}. The model was trained on grayscale images resized to $640 \times 640$ resolution with corresponding UGV bounding box annotations.

\subsection{Architecture and Training Details}

The YOLOv5 framework consists of three major components: a backbone for feature extraction, a neck for feature aggregation using PANet, and a head for predicting bounding boxes and class probabilities \cite{redmon2018, chowdhury2024}. The detection layers were customized for single-class (UGV) identification, and the top layers were adapted to accommodate grayscale inputs.
Training was conducted for 100 epochs using the Adam optimizer with a learning rate of 0.001 and batch size of 32. We monitored both classification loss and localization loss to evaluate training convergence.

\begin{figure}[htb]
    \centering
    \includegraphics[width=0.5\textwidth]{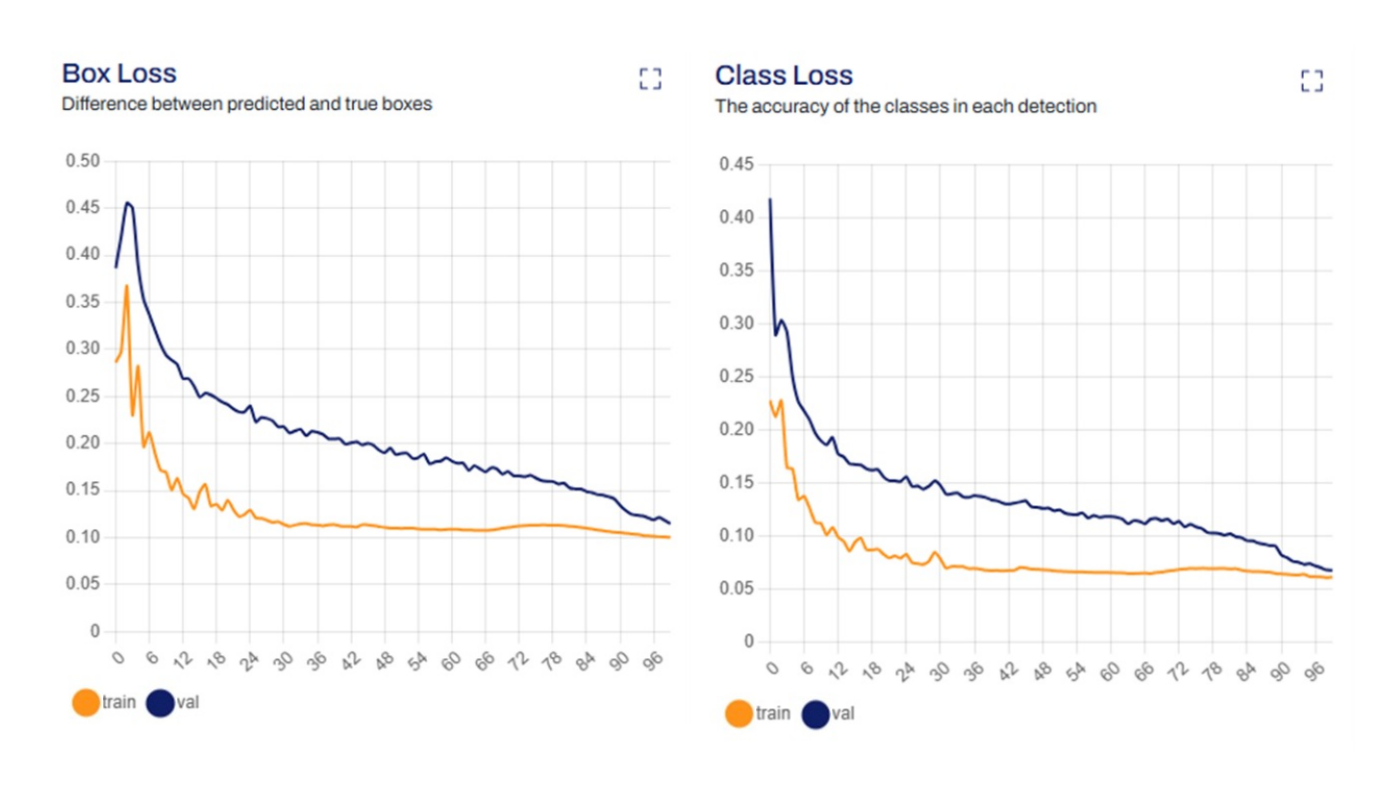}
    \caption{Training and validation loss curves: (Left) Box loss representing localization error. (Right) Class loss representing classification accuracy over epochs.}
    \label{fig:loss_curves}
\end{figure}
Figure~\ref{fig:loss_curves} shows the training progress. The box loss rapidly declined and stabilized, indicating improved bounding box prediction. The class loss also converged to a low value, suggesting strong detection reliability across training and validation sets.

The model achieved a mean average precision (mAP@0.5) of 95\%, computed using the standard Average Precision (AP) metric:

\[
AP = \sum_{n} (R_n - R_{n-1}) P_n
\]

where \( P_n \) is the precision at recall level \( R_n \). Since the system detects a single class (UGV), mAP and AP are equivalent in this context. The high mAP confirms the reliability of YOLOv5’s bounding box outputs, ensuring accurate UGV localization for downstream heading angle prediction.

\subsection{Evaluating YOLOv5 Bounding Box Accuracy for Heading Angle Calculation}
The system’s performance was first evaluated using YOLOv5-based UGV detection in combination with heading angle calculations derived from bounding box coordinates and VICON positional data. This setup establishes the baseline against which our neural network-based heading prediction model is later compared.

\begin{figure}[H]
    \centering
    \includegraphics[width=1\columnwidth]{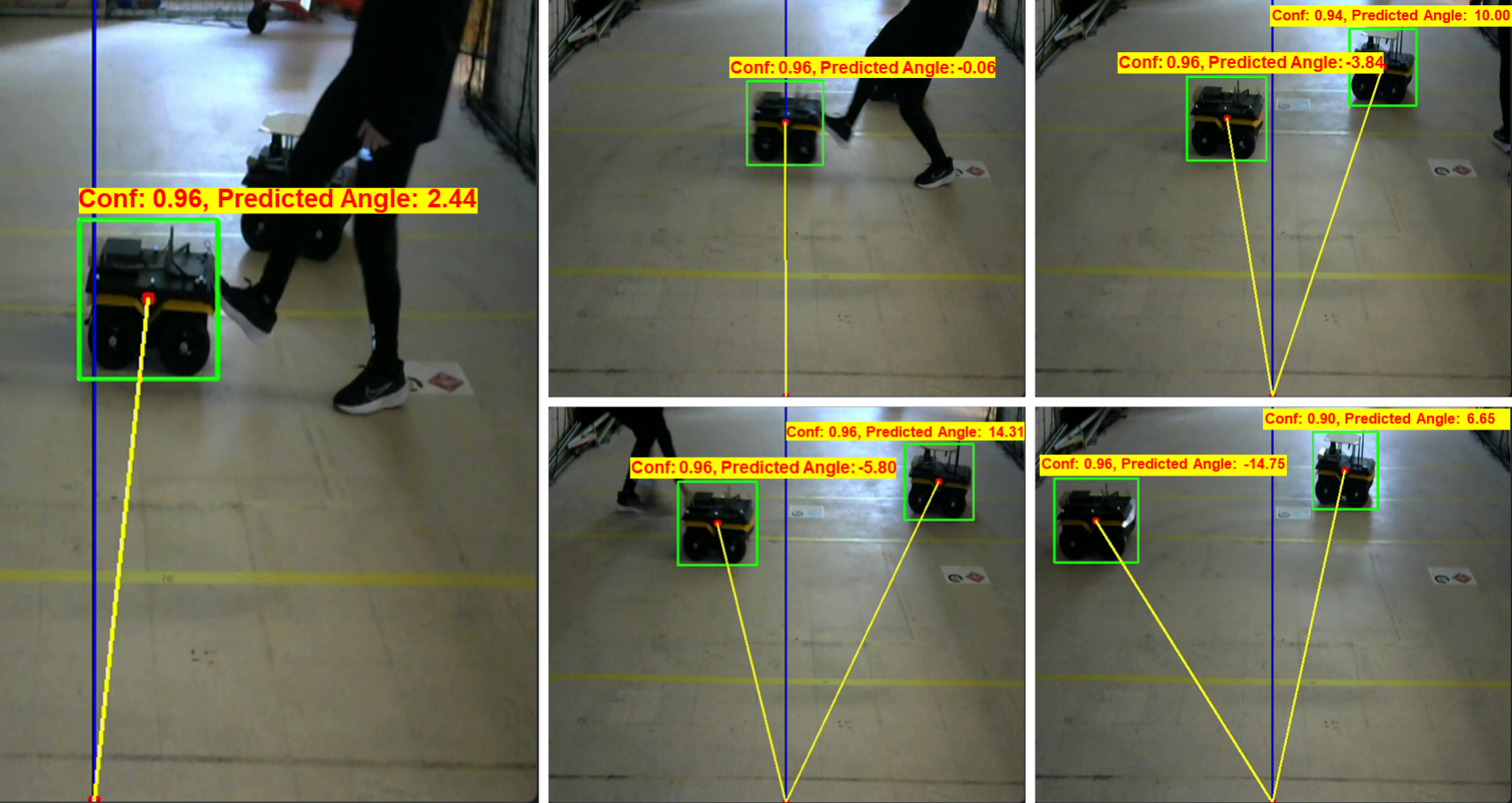}
    \caption{YOLOv5 UGV detection results with bounding boxes and corresponding heading angle calculations $\theta$. Green boxes indicate detected UGVs; yellow lines show angles computed using VICON and YOLO data.}
    \label{fig:yolo_heading_angle}
\end{figure}

Figure~\ref{fig:yolo_heading_angle} illustrates real-time UGV detection using YOLOv5. The bounding boxes (green) correspond to confidently detected UGVs, while the yellow lines represent calculated heading angles $\theta$ between the UAV and UGV. These angles were computed using VICON-derived positions of both agents and matched against the location of the bounding box center in the image frame.

\begin{figure*}[htb]
    \centering
    \includegraphics[width=\textwidth]{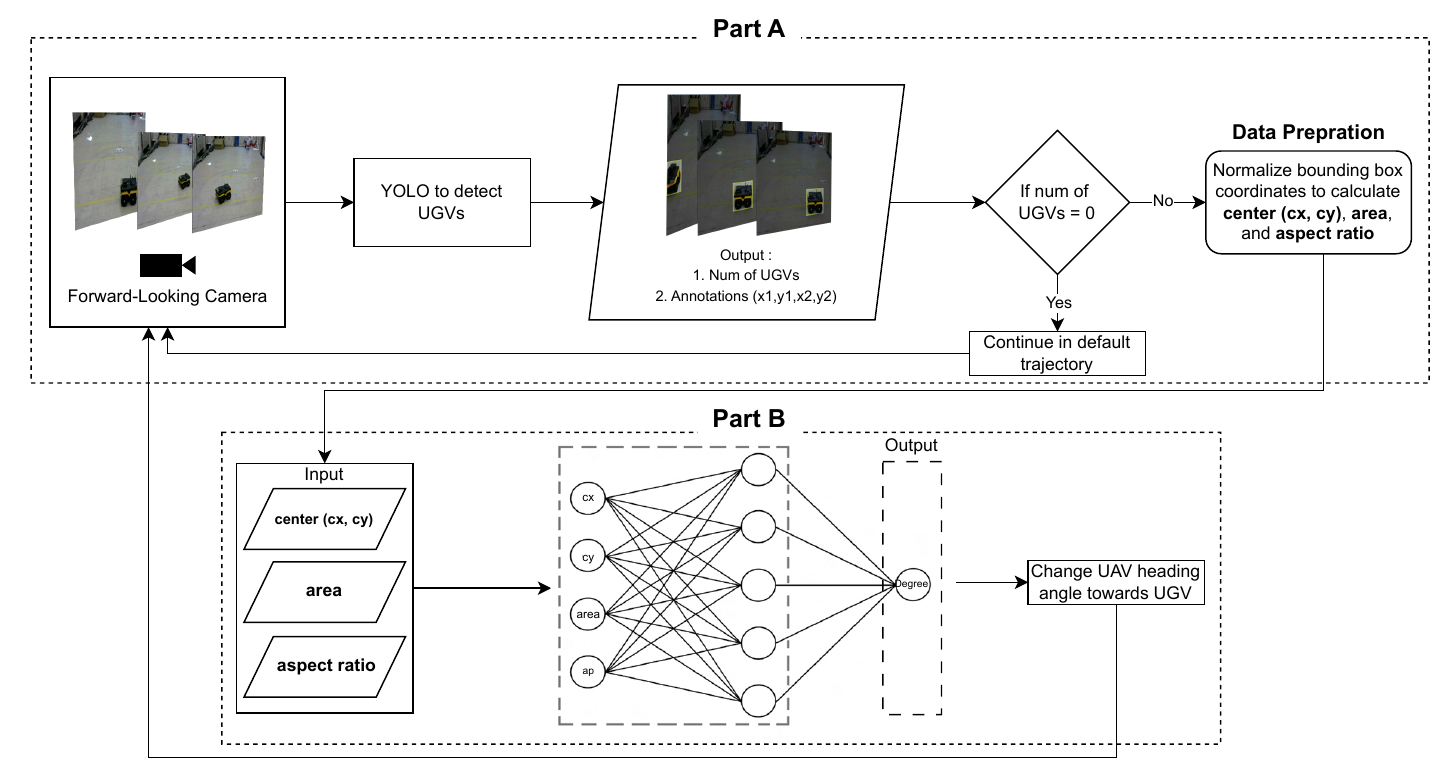}
    \caption{Overview of the heading angle prediction framework. \textbf{Part A}: YOLO detects UGVs and extracts bounding box features. \textbf{Part B}: The UAV sends normalized features to the ANN, which predicts the heading angle $\theta$.}
    \label{fig:ann_ugv_framework}
\end{figure*}

This approach demonstrated strong accuracy in spatial alignment, with a mean average precision of 95\% and minimal false positives. The heading angle calculations remained consistent across different UGV positions and trajectories, showcasing the reliability of YOLO bounding box outputs in guiding UAV orientation.

While effective, this method requires access to VICON data at runtime, limiting real-world applicability. Therefore, in the next section, we present a vision-only neural network model trained to replicate these heading angle predictions using only YOLO-extracted features, eliminating the need for an external localization infrastructure.

\section{ANN-Based Heading Angle Prediction}

To eliminate the dependency on external localization systems such as VICON during deployment, we developed a compact ANN model that predicts the heading angle $\theta$ based solely on visual features extracted from YOLOv5 bounding boxes. This vision-only approach allows real-time heading estimation using camera input, enabling scalable and autonomous UAV-UGV coordination in practical environments.

\subsection{ANN Architecture and Framework Design}

The ANN is designed as a lightweight feed-forward network with two hidden layers and ReLU activations. This relatively simple architecture was intentionally chosen because the input features ($c_x, c_y, A, \alpha$) already encode high-level spatial relationships, making deeper architectures unnecessary. Moreover, the lightweight design ensures fast inference suitable for real-time deployment on embedded hardware while accurately mapping the input features to a continuous-valued heading angle $\theta$.

\begin{itemize}
    \item Input: 4 normalized features ($c_x$, $c_y$, area, aspect ratio)
    \item Hidden Layers: 64 and 32 nodes with ReLU
    \item Output: 1 node predicting $\theta$
\end{itemize}

To provide a clearer understanding of the ANN model’s operational flow, Figure~\ref{fig:ann_ugv_framework} illustrates the overall framework. The system operates as a continuous two-stage pipeline, where the output of \textbf{Part A} directly serves as the input to \textbf{Part B}.

\begin{itemize}
    \item \textbf{Part A (YOLOv5 UGV Detection):} YOLOv5 first detects UGVs in the forward-looking camera feed. If no UGVs are detected, the UAV continues on its default trajectory. When UGVs are detected, YOLO outputs bounding box coordinates $(x_1, y_1, x_2, y_2)$, which are normalized to derive the center $(c_x, c_y)$, area, and aspect ratio. These features encode the relative position and size of the UGV in the UAV’s field of view.
    
    \item \textbf{Part B (ANN Heading Angle Prediction):} These extracted features are immediately fed into the ANN, which predicts the optimal heading angle $\theta$. The UAV continuously updates its orientation based on this prediction, aligning itself with the UGV for navigation or landing.
\end{itemize}
This sequential processing ensures that both modules complement each other: Part A provides real-time spatial perception, and Part B translates it into actionable heading control, as illustrated in Figure~\ref{fig:ann_ugv_framework}.

\subsection{Evaluation Methodology}

To validate the ANN’s accuracy, we compared predicted heading angles $\hat{\theta}$ against calculated ground truth values $\theta_{\text{true}}$ obtained from VICON data. The reference angle was defined as:

\begin{equation}
\theta_{true} = \text{atan2}(y_{\text{UGV}} - y_{\text{UAV}},\ x_{\text{UGV}} - x_{\text{UAV}})
\end{equation}

The following error metrics were used to assess model performance:
\begin{itemize}
    \item \textbf{Mean Absolute Error (MAE):}
    \[
    MAE = \frac{1}{N} \sum_{i=1}^{N} \left| \hat{\theta_{i}} - \theta_{\text{true}, i} \right|
    \]

    \item \textbf{Mean Squared Error (MSE):}
    \[
    MSE = \frac{1}{N} \sum_{i=1}^{N} \left( \hat{\theta_{i}} - \theta_{\text{true}, i} \right)^2
    \]

    \item \textbf{Root Mean Squared Error (RMSE):}
    \[
    RMSE = \sqrt{ \frac{1}{N} \sum_{i=1}^{N} \left( \hat{\theta_{i}} - \theta_{\text{true}, i} \right)^2 }
    \]
\end{itemize}

\section{Results and Discussion}\label{sec:resultfirst}

This section presents the performance results of the vision-based heading prediction pipeline. The ANN was trained for 100 epochs using the Adam optimizer (learning rate = 0.001, batch size = 32) and the mean squared error (MSE) loss. The final training loss was 0.0323, indicating stable convergence. Adam was chosen due to its adaptive learning rate capabilities and fast convergence, which are advantageous in real-time regression tasks involving low-dimensional feature spaces.

\begin{table}[htb]
    \centering
    \caption{Performance Metrics for ANN-Based Heading Angle Prediction}
    \label{tab:ann_performance}
    \begin{tabular}{|c|c|c|c|}
        \hline
        \textbf{Dataset} & \textbf{MSE} & \textbf{MAE} & \textbf{RMSE} \\
        \hline
        Validation & 0.0377 & 0.1446 & 0.1955 \\
        Testing    & 0.0383 & 0.1506 & 0.1957 \\
        \hline
    \end{tabular}
\end{table}

\begin{figure}[htb]
    \centering
    \includegraphics[width=1\columnwidth]{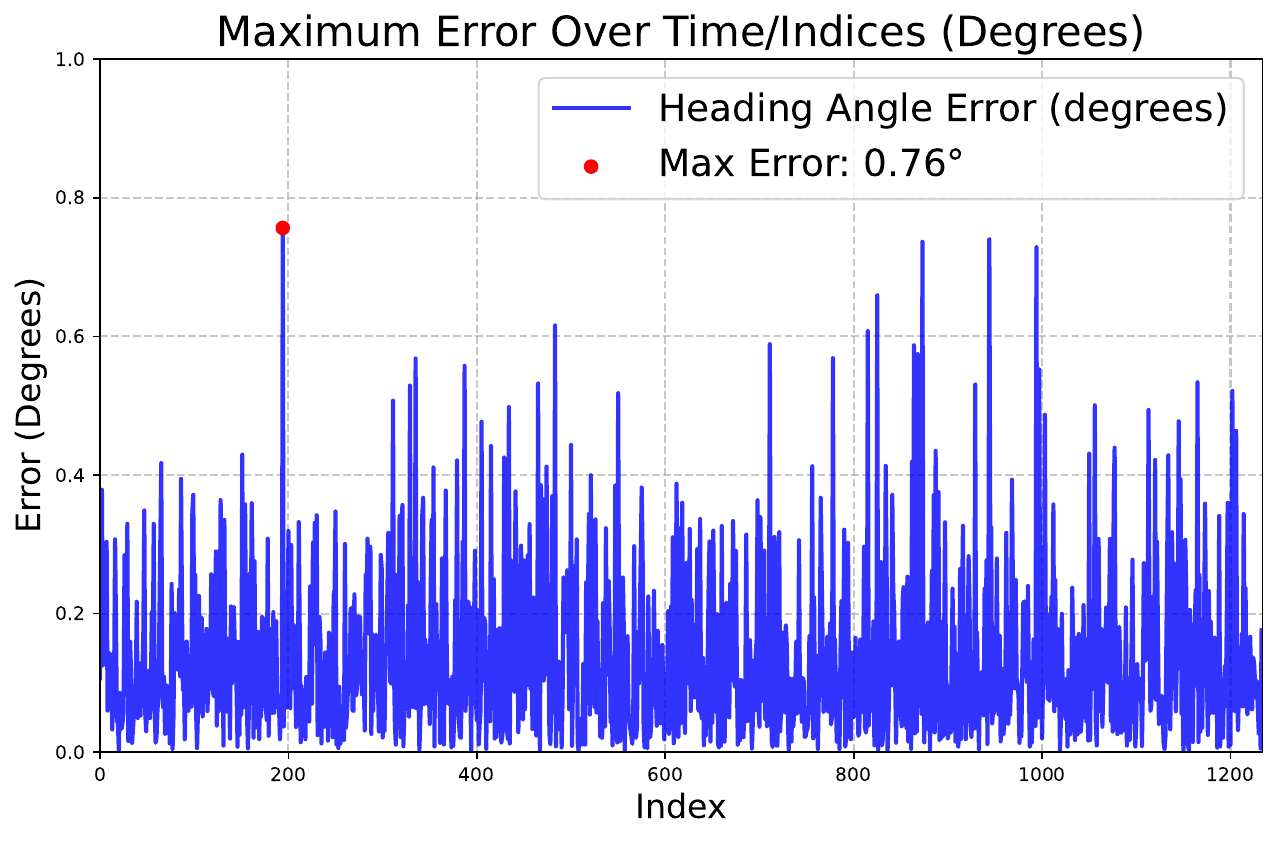}
    \caption{Absolute heading angle error across test set. The maximum observed error is 0.76°.}
    \label{fig:max_error_plot}
\end{figure}

Table~\ref{tab:ann_performance} summarizes the model's quantitative performance. The ANN demonstrated strong generalization, achieving a mean absolute error (MAE) of 0.1506° and a root mean squared error (RMSE) of 0.1957° on the test dataset. Figure~\ref{fig:max_error_plot} further illustrates the prediction reliability by plotting the absolute heading angle error across test indices.

\begin{figure}[htb]
    \centering
    \includegraphics[width=1\columnwidth]{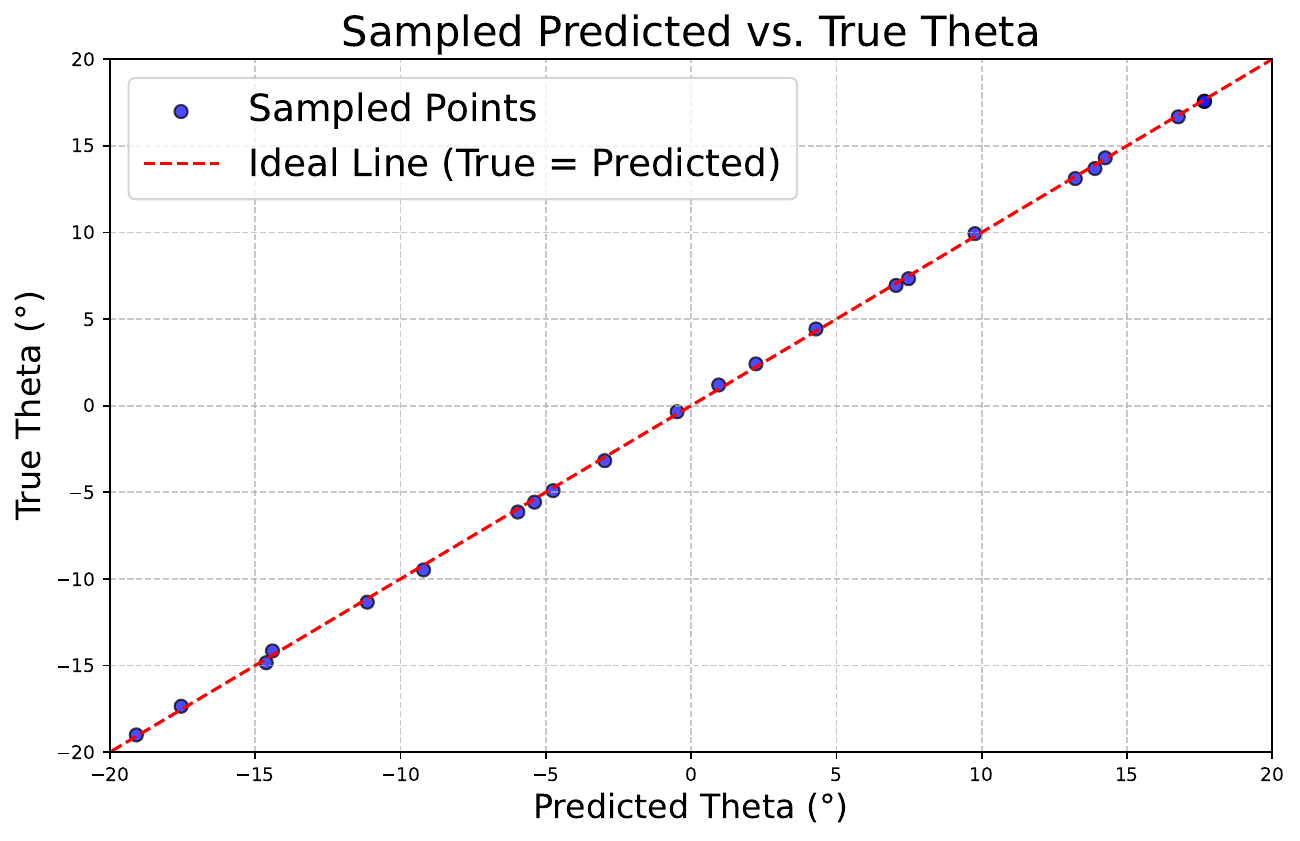}
    \caption{Representative scatter plot of predicted versus true heading angles. Points lie close to the dashed ideal line $\hat{\theta} = \theta_{\text{true}}$, indicating high prediction accuracy.}
    \label{fig:scatter_plot}
\end{figure}

The model maintained consistent performance throughout the sequence, with maximum observed errors of less than 1°, highlighting its robustness and suitability for real-time applications. Figure~\ref{fig:scatter_plot} presents a representative sample of predicted versus ground-truth heading angles. Even across challenging angular regions, the predicted values remain closely aligned with the true labels, further validating the model’s accuracy and its effectiveness for UAV-UGV coordination.

\begin{figure}[htb]
    \centering
    \includegraphics[width=1\columnwidth]{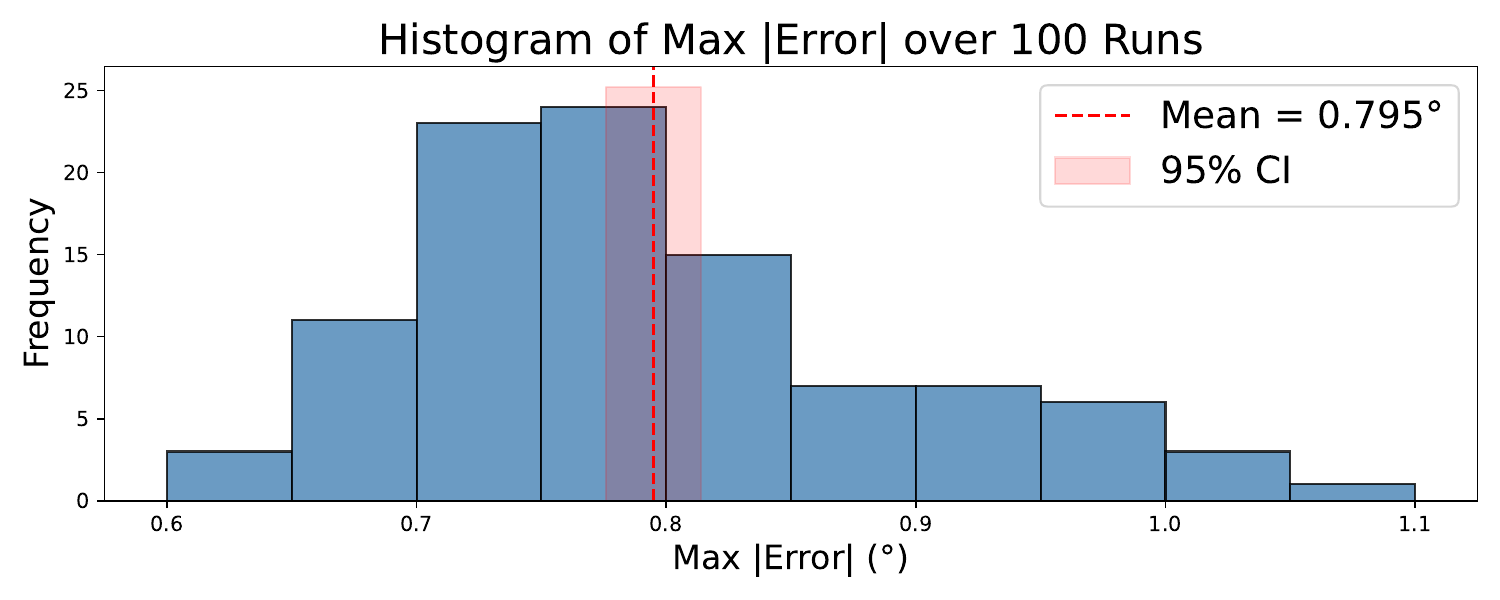}
    \caption{Distribution of maximum absolute heading-angle error (MaxAE) over 100 independent runs. The dashed line marks the mean (0.795°) with a shaded 95\% confidence interval (±0.019°). Ninety-six percent of runs achieved a MaxAE below 1°, indicating high consistency across random seeds.}
    \label{fig:max_error_histogram}
\end{figure}

To assess robustness to random initialization and minibatch ordering, we trained the model 100 times with different random seeds while keeping hyperparameters fixed (100 epochs, Adam optimizer, learning rate = 0.001, batch size = 32). For each run, we evaluated the maximum absolute error (MaxAE) on a fixed test set. As show in Figure~\ref{fig:max_error_histogram}, the distribution of MaxAE values was tightly concentrated around a mean of 0.795° (95\% CI: ±0.019°; std = 0.097°), demonstrating low run-to-run variability. Notably, 96\% of runs achieved a MaxAE below 1°, underscoring consistency across seeds. Complementary metrics further confirmed sub-degree accuracy: MAE = 0.119° ± 0.003° (std = 0.014°) and RMSE = 0.162° ± 0.003° (std = 0.013°). These results validate the model’s reliability in both typical and worst-case scenarios.

In addition to accuracy, latency and detection reliability are critical for real-time autonomous navigation. Therefore, we evaluated the computational performance of the integrated pipeline on embedded hardware to assess its feasibility for time-sensitive deployment scenarios. The following results demonstrate the system’s real-time readiness and detection precision.
Combined inference latency for the YOLOv5 detector and ANN predictor averaged 31 ms per frame, enabling real-time execution at 33–35 FPS on embedded hardware. The YOLOv5 model also achieved a mean average precision of 95\%, ensuring reliable UGV detection. These results validate the proposed pipeline's suitability for real-time heading angle prediction in GPS- and GNSS-denied environments without reliance on external localization.
\subsection{Qualitative Results}

To qualitatively assess the performance of the ANN-based heading angle prediction model, we compared its outputs against the ground truth angles calculated using VICON data. This section presents visual results across three different stages of UGV movement to illustrate the model’s real-time response and generalization capabilities.
\begin{figure}[h]
    \centering
    \includegraphics[width=0.93\columnwidth]{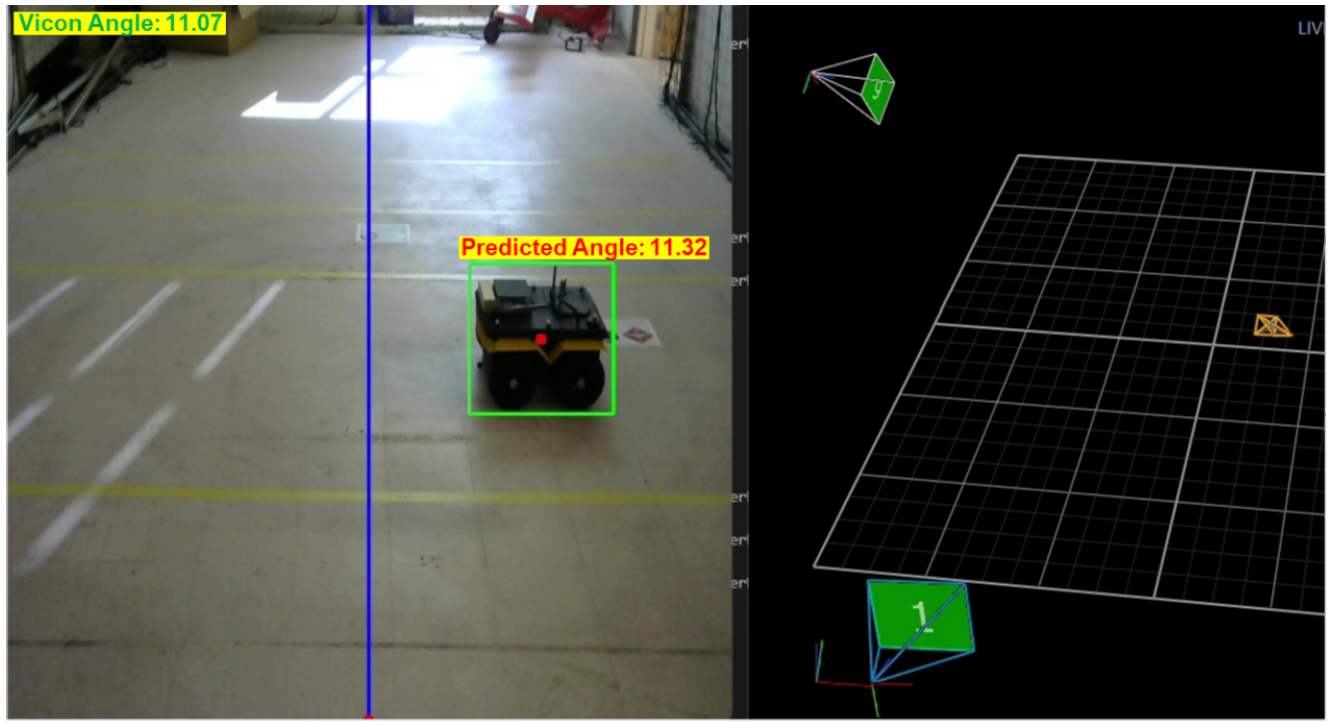}
    \caption{Stage 1: Comparison of ANN-predicted heading angle (red) with ground truth angle (green) as the UGV enters the scene from the right.}
    \label{fig:stage1}
\end{figure}

\begin{figure}[h]
    \centering
    \includegraphics[width=0.93\columnwidth]{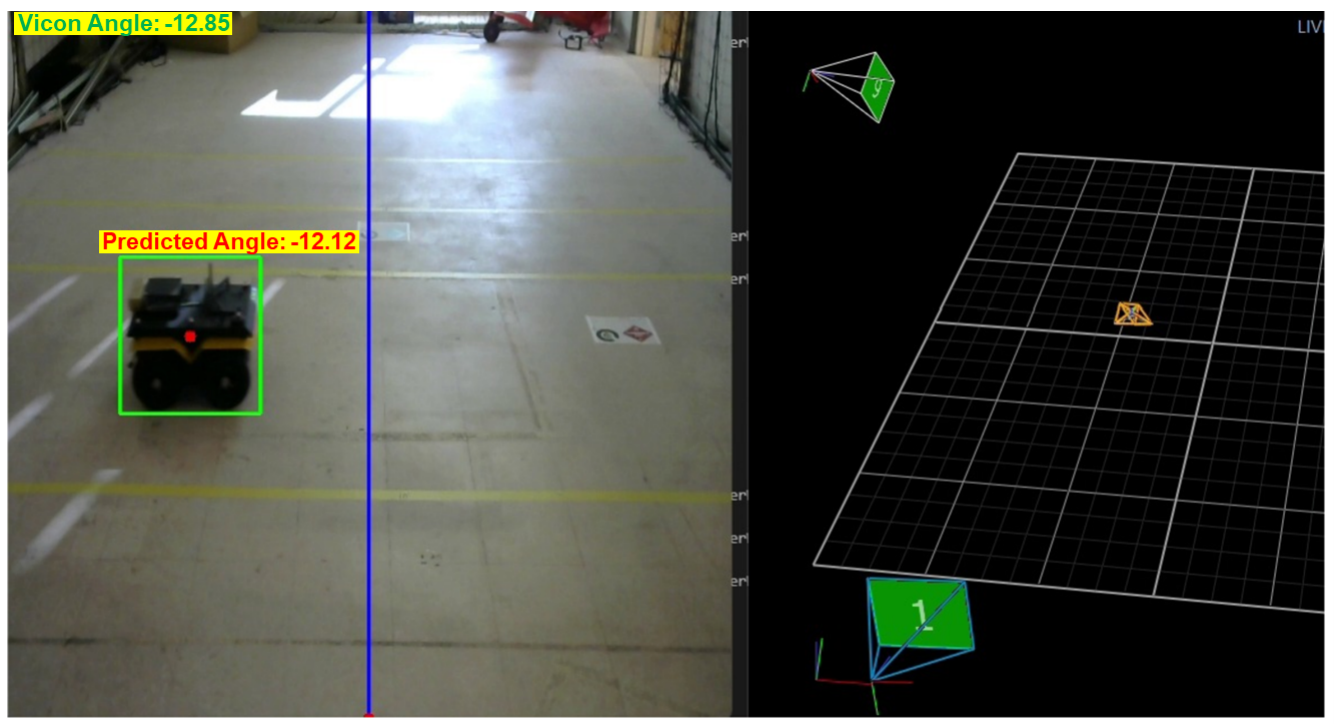}
    \caption{Stage 2: Heading angle prediction during mid-range UGV movement. The ANN closely follows the calculated reference.}
    \label{fig:stage2}
\end{figure}

\begin{figure}[!h]
    \centering
    \includegraphics[width=0.93\columnwidth]{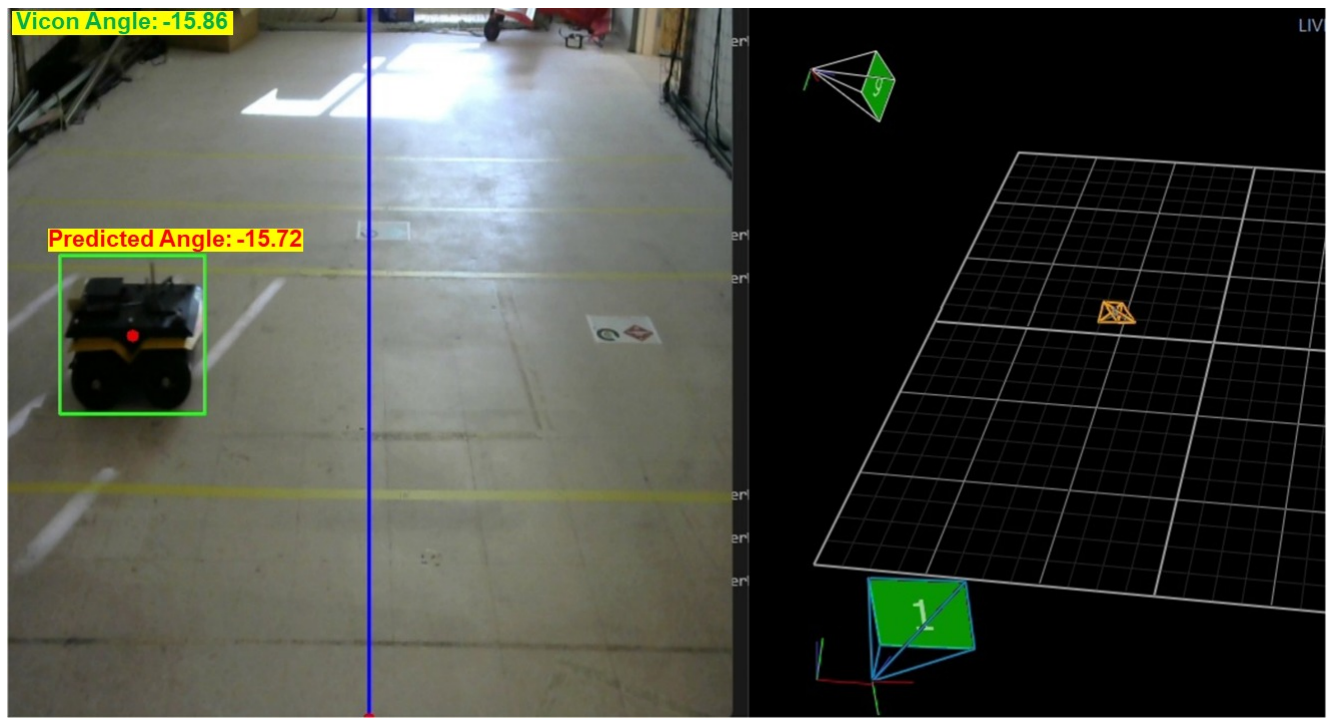}
    \caption{Stage 3: Final stage as the UGV approaches end of the scene. ANN-predicted angle remains consistent with ground truth.}
    \label{fig:stage3}
\end{figure}

Figures~\ref{fig:stage1}–\ref{fig:stage3} show the heading angle visualizations over time. The green text represent the calculated ground truth heading angles ($\theta_{\text{true}}$) derived from VICON, while the red text indicate the ANN predictions ($\hat{\theta}$). The overlap between predicted and actual angles highlights the ANN’s ability to track dynamic UGV positions without access to positional ground truth during inference.

The ANN consistently maintained low error across varying UGV trajectories, distances, and bounding box sizes, demonstrating robustness in the presence of visual changes. These visual results reinforce the quantitative findings from Section~\ref{sec:resultfirst} and confirm that the ANN model performs accurately under dynamic conditions.

\begin{figure*}[!htbp]
    \centering
    \captionsetup{justification=centering}
    \includegraphics[width=\textwidth]{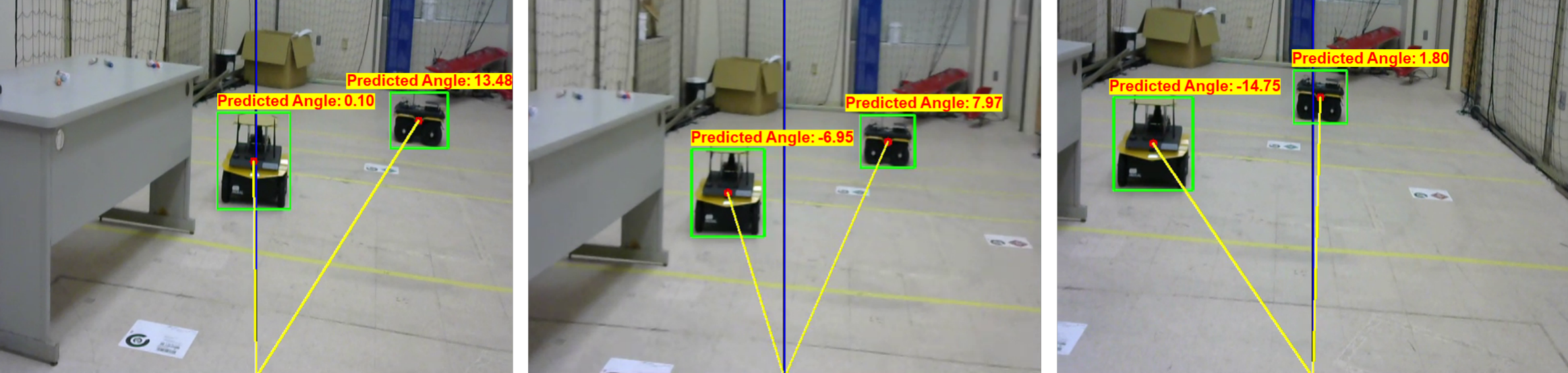}
    \caption{Visualization of predicted heading angles from the UAV’s perspective. Each frame shows the predicted heading angle (in red) overlaid on the YOLOv5-detected UGV bounding box. The yellow lines represent the predicted direction toward the UGV relative to the UAV’s forward-facing camera axis (blue line), simulating real-time aerial orientation adjustment.}
    \label{fig:comparison_extra1}
\end{figure*}

To further support this, additional qualitative examples are shown in Figure~\ref{fig:comparison_extra1}, covering a broader range of viewpoints, distances, and orientations. In these cases, we simulate realistic UAV operating conditions where the drone is in motion and must dynamically predict the required heading angle toward multiple UGV targets. This setting more closely reflects practical scenarios in which the UAV actively adjusts its orientation during mission execution.

\subsection{Comparative Results}
\label{sec:comparative_results}

To place our results in context, we compare the proposed lightweight, markerless, vision-only heading framework with a representative prior study for UAV-assisted UGV detection and following that we refer to as \emph{Husky Detection \& Tracking} \cite{amil2024vision}. The Husky system integrates YOLOv7 object detection, ByteTrack multi-object association, and ArUco-marker–based relative pose estimation within a multi-sensor outdoor platform instrumented with LiDAR, IMU, and GPS; Therefore Husky relies on Aruco markers, sensor fusion, and extensive camera calibration for outdoor UGV tracking and following. In contrast, our approach targets rapid deployment in GPS-denied or infrastructure-sparse environments: a fine-tuned single-class YOLOv5 detector feeds compact bounding-box features to a lightweight feed-forward ANN that directly regresses the required UAV heading angle, eliminating the need for external fiducials, calibrated camera extrinsics at runtime, or global positioning infrastructure.

\newcommand{\oursGPU}{RTX\,3060\xspace}           
\newcommand{\oursCPU}{Intel\,Core\,i7-12700K\xspace}
\newcommand{\oursRAM}{32\,GB\xspace}
\newcommand{\oursYOLOSize}{15\,MB\xspace}
\newcommand{\oursANNSize}{$<\,$2\,MB\xspace}

\newcommand{\oursFullMS}{43\xspace}    
\newcommand{\oursFPSObs}{33--35\xspace} 
\newcommand{\oursInferMS}{31\xspace}

\subsubsection{Systems at a Glance}
Husky Detection \& Tracking \cite{amil2024vision} addresses outdoor off-road UGV operations by combining a YOLOv7 detector, ByteTrack temporal association, and ArUco-marker--based pose estimation on a Husky A200 ground vehicle instrumented with LiDAR, IMU, and GPS. Our system focuses instead on \emph{relative heading alignment} in GPS-denied or infrastructure-sparse environments using only monocular vision: YOLOv5 provides single-class UGV detections whose bounding-box features drive a compact ANN regressor that outputs the required UAV heading angle.

\subsubsection{Quantitative Accuracy}
Our ANN achieves 0.1506$^{\circ}$ mean absolute error (MAE), 0.1957$^{\circ}$ RMSE, and a maximum error of less than 1$^{\circ}$.

The Husky paper reports qualitative and tabulated yaw discrepancies derived from ArUco pose estimates under different view angles (e.g., 15$^{\circ}$ true $\rightarrow$ 14$^{\circ}$ est; 30$^{\circ}$ true $\rightarrow$ 25$^{\circ}$ est), implying errors up to roughly 12$^{\circ}$ in challenging viewpoints; an aggregate MAE is not provided.

\subsubsection{Runtime and Model Footprint}
We measured YOLOv5+ANN inference at \oursInferMS{}~ms per frame on a \oursGPU{} GPU with an \oursCPU{} host and \oursRAM{} RAM. The combined model footprint is under 17~MB (\oursYOLOSize{} detector weights + \oursANNSize{} ANN).

The Husky study describes a mini-ITX compute box with an Nvidia GPU but does not report detector/tracker model sizes, VRAM usage, or per-frame latency, preventing direct quantitative scaling comparisons. However, based on the fact that their detector is derived from YOLOv7, it can be estimated to be significantly heavier than a comparable YOLOv5 model, YOLOv5 has approximately 7.2M parameters ($\sim$14\,MB), whereas YOLOv7 has around 36.9M parameters ($\sim$71\,MB), making it roughly five times larger in model size and computational demand.

\subsubsection{Sensor and Infrastructure Dependencies}
Our runtime stack requires only an onboard monocular camera to estimate heading angles directly from visual bounding box features, eliminating the need for external infrastructure. All motion-capture supervision is confined to offline training.

Husky depends on the visibility of ArUco markers for accurate pose recovery and was developed on a platform equipped with LiDAR, IMU, and GPS. However, this approach depends heavily on fiducial markers and additional sensor calibration, making it less suitable for rapid deployment in unstructured environments and increasing the setup burden in austere conditions.

Table~\ref{tab:comparison_ours_husky} summarizes the key algorithmic, performance, and deployment trade-offs between the proposed markerless heading framework and Husky Detection \& Tracking.

\begin{table}[!htbp]
    \centering
    \caption{Comparison between the proposed method and Husky Detection \& Tracking. Bold entries indicate the approach that is advantageous for minimally instrumented, GPS-denied deployment scenarios.}
    \label{tab:comparison_ours_husky}
    \renewcommand{\arraystretch}{1.25}
    \large
    \resizebox{\columnwidth}{!}{%
    \begin{tabular}{|p{3.0cm}|p{5.0cm}|p{5.0cm}|}
        \hline
        \textbf{Criteria} & \textbf{Ours} & \textbf{Husky~\cite{amil2024vision}} \\ \hline

        \textbf{Heading Error} &
        \textbf{0.1506$^{\circ}$ MAE; Less than 1$^{\circ}$} &
        Yaw examples: 15$^{\circ}$ true $\rightarrow$ 14$^{\circ}$ est; 30$^{\circ}$ true $\rightarrow$ 25$^{\circ}$ est (up to $\sim$12$^{\circ}$). no aggregate MAE reported \\ \hline

        \textbf{Hardware} &
        \textbf{\oursGPU{} + \oursCPU{} + \oursRAM{} RAM} &
        mini-ITX computer w/ ``GTX 1065'' GPU \\ \hline

        \textbf{Inference Latency} &
        \textbf{\oursInferMS{} ms / frame} &
        N/R \\ \hline

        \textbf{Throughput (FPS)} &
        \textbf{\oursFPSObs{} FPS} &
        N/R \\ \hline

        \textbf{Sensor Dependency} &
        \textbf{Monocular camera only} &
        ArUco + calibrated camera; platform includes IMU, GPS \\ \hline

        \textbf{Tracking Modality} &
        \textbf{Heading regression per frame} &
        ByteTrack + Kalman \\ \hline

        \textbf{Pose / Heading Source} &
        \textbf{ANN regression from bounding-box geometry (markerless)} &
        ArUco pose solve from calibrated camera; Euler conversion for yaw \\ \hline

        \textbf{Marker Dependency} &
        \textbf{None} &
        Requires visible ArUco tags on UGV \\ \hline

        \textbf{Dataset Scale} &
        \textbf{9{,}000 images total} &
        1{,}058 images total \\ \hline

        \textbf{Model Size} &
        \textbf{YOLOv5 $\sim$7.2M Parameters ($\sim$14MB)} &
        YOLOv7 $\sim$36.9M Parameters ($\sim$71MB) \\ \hline

        \textbf{Multiple Vehicle Detection \& Tracking} &
        \textbf{Multiple UGV} &
        Single UGV \\ \hline
        
        \textbf{Computational Footprint} &
        \textbf{Lightweight single-class YOLOv5 + small ANN ($\approx$17 MB).} &
        Heavier stack: YOLOv7 + ByteTrack + marker detection/pose; no size/runtime data reported \\ \hline

        \textbf{Primary Limitations} &
        Indoor training data &
        Marker occlusion; yaw error growth at large angles; small dataset; no runtime/model footprint reporting. \\ \hline
    \end{tabular}}
\end{table}

\begin{figure}[htb]
    \centering
    \includegraphics[width=1\columnwidth]{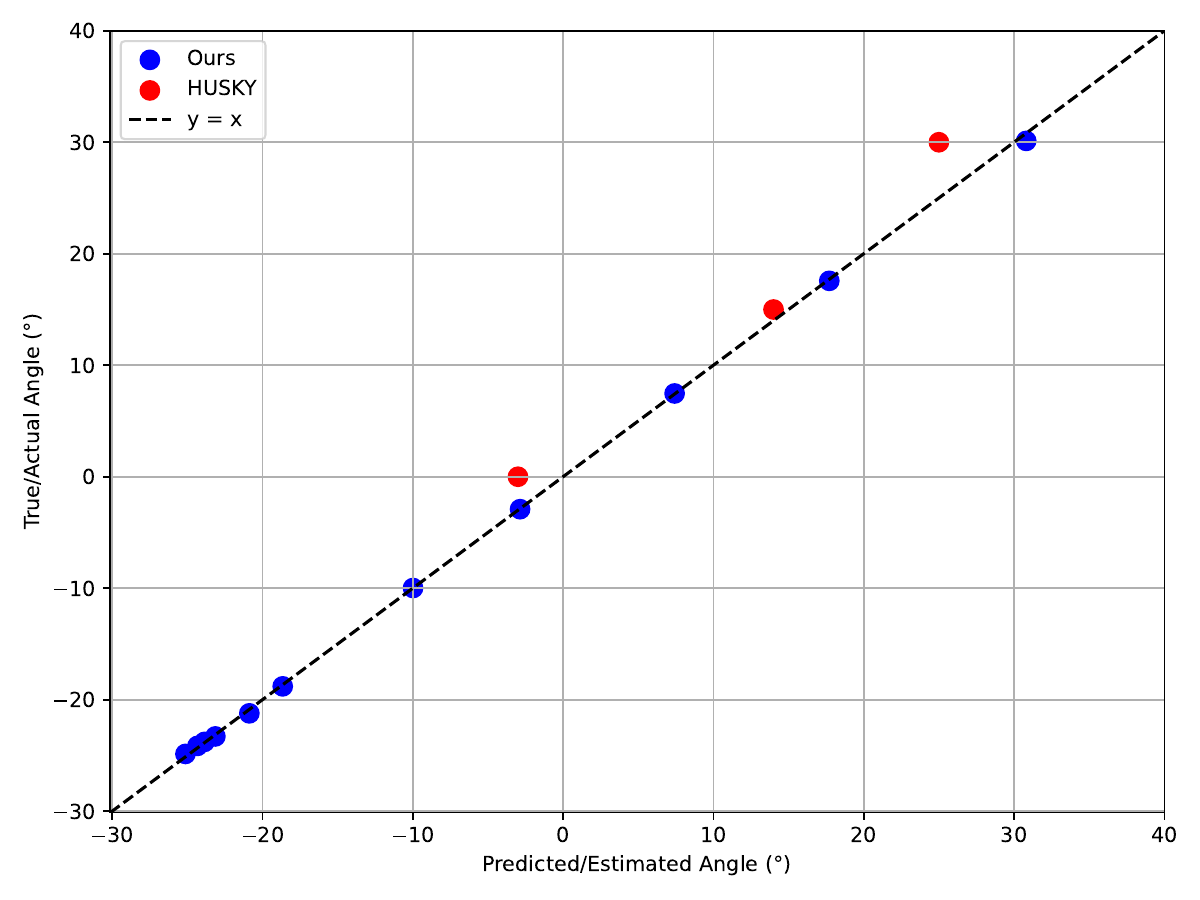} 
    \caption{Comparison of predicted vs. true heading angles between our approach (blue) and Husky’s Aruco-based method (red). The dashed line represents perfect alignment ($y = x$).}
    \label{fig:comparison_husky}
\end{figure}

Figure~\ref{fig:comparison_husky} then plots predicted versus ground-truth heading: the blue trace shows our ANN-based regression results, while the red points indicate representative Husky ArUco-based yaw estimates reported in \cite{amil2024vision}. The tight clustering of our predictions about the $y=x$ line highlights the sub-degree accuracy achieved by the proposed approach.

In summary, the proposed framework complements Husky Detection \& Tracking by solving the relative‐heading alignment problem with a minimal, markerless, monocular pipeline that runs on embedded hardware and requires no external infrastructure. Husky demonstrates strong robustness for long‑duration outdoor UGV following when calibrated ArUco fiducials and multi‑sensor fusion (LiDAR/IMU/GPS) are available, but these dependencies increase deployment cost and complexity. Our approach is therefore well suited to infrastructure‐less, GPS‑denied operations that demand quick setup, small model footprint, and sub‑degree heading precision.

\section{System-Level Evaluation}

The final evaluation assessed the complete system operating under real-time conditions, where YOLOv5 performed UGV detection, the ANN predicted the relative heading angle, and the UAV continuously adjusted its orientation based on the predicted heading. A trial was considered successful if the UAV maintained a heading alignment error below 1° for the majority of the approach duration and completed the alignment sequence without manual intervention.

The system achieved a 95\% success rate under these conditions. The mean absolute heading alignment error during closed-loop operation was 0.15°, with a maximum observed error of less than 1°, consistent with the offline ANN evaluation. These results demonstrate that the integrated YOLO-ANN pipeline can reliably guide the UAV toward the UGV without reliance on external localization systems.

\section{Conclusion}

This paper presented a vision-based framework for real-time UAV-UGV coordination, designed to function without reliance on external localization systems such as GPS, GNSS, or motion capture. By integrating a fine-tuned YOLOv5 model for UGV detection with a lightweight ANN for heading angle prediction, the system achieves reliable aerial-ground alignment using only monocular camera input.

The end-to-end pipeline operates within a ROS-based control loop and demonstrates strong performance, with a mean absolute heading error of 0.1506°, an mAP of 95\%, and a combined average inference time of 31 milliseconds per frame. These results confirm the system’s suitability for dynamic, real-time operations in environments where localization infrastructure is limited or unavailable.

A key contribution of this work lies in demonstrating the potential of vision-only frameworks for GPS and GNSS-denied or degraded environments. While this study was conducted in a controlled laboratory setting, the results indicate promising applicability to indoor, subterranean, or contested environments, as well as to unstructured terrains where learning-based perception methods are increasingly required for robust navigation \cite{guastella2020learning}. The proposed framework offers a cost-effective and lightweight solution with potential for future deployment in mission-critical scenarios where traditional localization methods are not feasible.

\section{Future Work}

Future efforts will focus on enhancing the system’s robustness for real-world deployment. This includes integrating depth estimation, lighting-invariant perception, and temporal filtering to improve performance under challenging conditions such as occlusion, outdoor variability, or low light \cite{gasperini2023robust}, alongside real-time adaptation to unfamiliar terrain through self-supervised learning techniques \cite{chen2023learning}.

We also plan to explore adversarial robustness strategies, such as adversarial training and anomaly detection, to safeguard against visual attacks and data manipulation. These measures are particularly important for safety-critical applications.

Although this work focuses on the functional capabilities of vision-based heading prediction, it is also important to consider potential vulnerabilities in such pipelines. In related research, we explored how Trojan attacks can manipulate deep learning models used for autonomous navigation \cite{trojan2023, rezaifstrojan}. These findings emphasize the need for future extensions of this system to include adversarial robustness and interpretability in critical applications.

This real-time evaluation confirms that the integrated YOLO-ANN pipeline can perform reliable heading estimation in a closed-loop control system without reliance on external sensors. Future work may focus on deploying the system in outdoor environments, extending perception capabilities, and enhancing security assurance.

\section*{Acknowledgments}
This project is primarily supported by the National Science Foundation under Grant No. 2301553 and the University Transportation Center (UTC) of the USA through Grant No. 69A3552348327. Additionally, partial support is provided by NASA-ULI under Cooperative Agreement No. 80NSSC20M0161.


\bibliographystyle{IEEEtran}
\bibliography{ref}


 




\vfill

\end{document}